\documentclass[acmlarge,screen]{acmart}

\setcopyright{none}
\copyrightyear{2021}
\acmYear{2021}
\acmDOI{}

\acmConference[FragileEarth '21]{Fragile Earth '21: Fragile Earth: Accelerating Progress towards Equitable Sustainability}{August 14--18, 2021}{Virtual}
\acmPrice{}
\acmISBN{}

\DeclareMathOperator{\boldx}{\mathbf{X}}
\DeclareMathOperator{\boldy}{\mathbf{Y}}

\definecolor{figred}{RGB}{184,84,80}
\definecolor{figblue}{RGB}{108,142,191}
\definecolor{figyellow}{RGB}{214,182,86}

\newcounter{comments}

\begin{document}

\title{Detecting Cattle and Elk in the Wild from Space}

\author{Caleb Robinson}
\authornote{Both authors contributed equally to this research.}
\email{caleb.robinson@microsoft.com}
\affiliation{%
  \institution{Microsoft, AI for Good Research Lab}
  \country{USA}
}

\author{Anthony Ortiz}
\authornotemark[1]
\email{anthony.ortiz@microsoft.com}
\affiliation{%
  \institution{Microsoft, AI for Good Research Lab}
  \country{USA}
}

\author{Lacey Hughey}
\affiliation{%
  \institution{Smithsonian Conservation Biology Institute}
  \country{USA}
}

\author{Jared A. Stabach}
\affiliation{%
  \institution{Smithsonian Conservation Biology Institute}
  \country{USA}
}

\author{Juan M. Lavista Ferres}
\affiliation{%
  \institution{Microsoft, AI for Good Research Lab}
  \country{USA}
}

\renewcommand{\shortauthors}{Robinson and Ortiz, et al.}

\begin{abstract}
Localizing and counting large ungulates -- hoofed mammals like cows and elk -- in very high-resolution satellite imagery is an important task for supporting ecological studies. Prior work has shown that this is feasible with deep learning based methods and sub-meter multi-spectral satellite imagery.
We extend this line of work by proposing a baseline method, CowNet, that simultaneously estimates the number of animals in an image (counts), as well as predicts their location at a pixel level (localizes).
We also propose an methodology for evaluating such models on counting and localization tasks across large scenes that takes the uncertainty of noisy labels and the information needed by stakeholders in ecological monitoring tasks into account.
Finally, we benchmark our baseline method with state of the art vision methods for counting objects in scenes. We specifically test the temporal generalization of the resulting models over a large landscape in Point Reyes Seashore, CA.
We find that the LC-FCN model performs the best and achieves an average precision between $0.56$ and $0.61$ and an average recall between $0.78$ and $0.92$ over three held out test scenes.
\end{abstract}



\maketitle

\section{Introduction}
Over the past decade, very high-resolution (VHR) remotely sensed imagery has become increasingly available to conservation scientists for research purposes.
Processing these data manually, however, is a labor intensive task that requires specialized skills and equipment. Further, the effort required to process VHR imagery for landscape-scale ecological analyses is prohibitive. For example, it typically takes manual annotators weeks or months to complete labeling tasks required by biological assessments conducted at ecologically meaningful spatial scales. However, the ability to accurately assess environmental change across large, remote landscapes is essential to conservation success and has many applications in scientific research, policy making, and protected area management.

VHR imagery holds particular promise as a tool for conducting wildlife population surveys across a range of species and environmental contexts (e.g., \cite{larue2017feasibility,wang2019surveying}). Recent work, for example, demonstrates that VHR data can be used to study fundamental aspects of behavioral ecology, such as species interactions in ecologically sensitive landscapes~\cite{hughey2021effects} and emergent properties of collective motion in animal groups~\cite{hughey2018challenges}. The early promise of such studies underscores the need for efficient, scalable techniques to rapidly extract large amounts of information from VHR imagery.

\begin{figure*}[tb]
    \centering
    \includegraphics[width=1.0\linewidth]{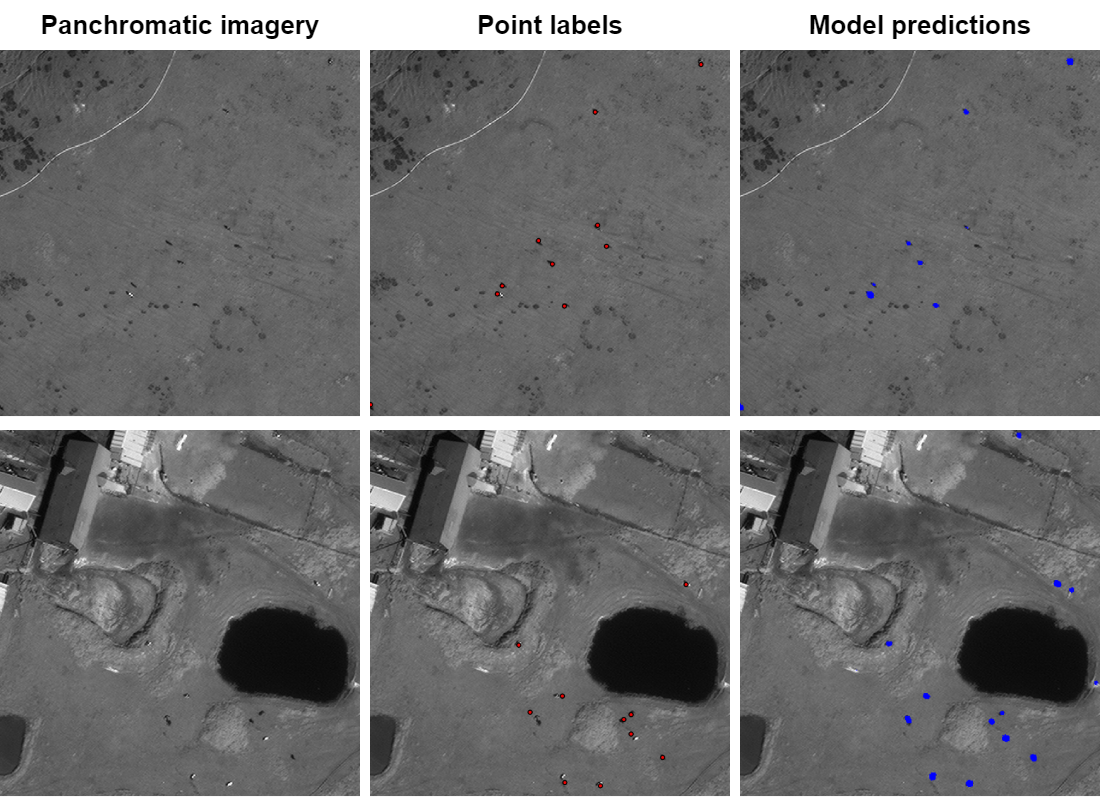}
    \caption{Examples of VHR (0.3m/px) panchromatic imagery, point labels, and CowNet model localization predictions for two areas in a held out test scene.}
    \label{fig:examples}
\end{figure*}

Automated approaches can bridge the gap between the amount of high-resolution imagery available for performing ecological analyses and the amount of manual labor required to interpret the imagery. For instance, deep learning approaches for counting and localizing cattle in high-resolution imagery (at spatial resolutions of $<0.5\text{ m}/\text{pixel}$) have recently been shown to be possible~\cite{laradji2020counting}. Other work using object detection models has been shown to detect and count African elephants, with a key goal of generalizing models across new landscapes where training data do not exist~\cite{duporge2020using}.
Fully automated approaches also exist, with researchers surveying seals on pack ice at a rate ten times faster than experienced human observers~\cite{gonccalves2020sealnet}.

At least two challenges exist in using automated methods for supporting ecological analyses with high-resolution remotely sensed imagery.  These include  (1) designing or adapting methods to work in new settings, either spatially or temporally, and (2) measuring the effectiveness of results with reliable validation data. For example, Hughey et al. measure the spatial overlap between cattle and reintroduced elk populations partially supported by animal location data derived from high-resolution satellite imagery over multiple points in time~\cite{hughey2021effects}. For an automated approach to adequately support this ecological task, models must \textit{localize} cattle and elk in large amounts of high-resolution imagery with high recall, and generalize across imagery taken at the same location for different points in time.

In this work, we aim to explore both of these challenges within the context of the aforementioned example -- counting and localizing cattle and elk in a time series of VHR imagery over Point Reyes, CA in the USA.
First, we propose a baseline deep learning-based approach for counting and localizing cows and elk. This approach treats counting and localizing animals as two separate tasks, each of which share the same spatial representation.
Second, we test the performance of automated methods on this problem in several novel ways: we explicitly account for the spatial noise in our labeled data to better estimate model performance and we propose metrics that can be applied over large scenes (as opposed to fixed patch based evaluations).
We explicitly test the temporal generalization of our baseline method and several other common methods from computer vision literature.

\section{Related Work}

Different approaches have been proposed for localizing and counting objects using computer vision techniques. The problem of counting objects have been approached in two different directions. The first approach is detection-based counting, which requires object detection or segmentation before performing the counting task~\cite{girshick2014rich,laradji2018blobs,laradji2020looc}. The second approach is density estimation-based and does not requires detection or segmentation to predict for the count~\cite{xie2018microscopy,li2018csrnet}.

The most related work to ours is by Laradji et al.~\cite{laradji2020counting} which compares CSRNet~\cite{li2018csrnet} and LC-FCN~\cite{laradji2018blobs} modeling approaches 
on the task of counting and localizing the number of cattle in high resolution satellite imagery using a dataset of 12,252 labelled patches from Maxar satellites.
The authors compare models with the mean absolute percentage error (MAPE) and grid average mean absolute percentage error (GAMPE) metrics on different groups of held out patches that are partitioned by density.
We significantly extend this evaluation methodology considering the questions of: ``How to use noisy point labels?'', ``How to evaluate model performance at a scene level?'', and ``How to evaluate counting and localization performance separately?''. This is important for using such models in ecological applications as we discuss later.

\section{Problem formulation}
We assume that we are given a time-series of $t$ VHR satellite image scenes covering roughly the same spatial location, $\left[\mathbf{X}^1, \ldots, \mathbf{X}^t \right]$, as well as noisy point labels for each scene that indicate the presence of a cow/elk. These point labels can be represented as per scene binary \textit{masks}, $\left[\mathbf{Y}^1, \ldots, \mathbf{Y}^t \right]$ where $\mathbf{Y}^i_{xy}$ is $1$ if there is a cow/elk within $d$ meters from the $y^\text{th}$ row and $x^\text{th}$ column of scene $i$, but is $0$ otherwise. The spatial dimensions of $\mathbf{Y}^i$ are the same as $\mathbf{X}^i$. Here, the value $d$ represents the extent of the spatial noise within the dataset and will often be unknown for datasets that have been curated without machine learning applications in mind\footnote{For example, it is unlikely that an ecologist will need pixel perfect point labels of the locations of cow and elk over large scenes and will tolerate errors of several meters when creating such datasets.}. Importantly, we do not assume that the time-series of scenes are precisely co-registered. For example, the scenes can be imaged from large off-nadir angles and have errors in their georegistration. We say that the scenes in the time series cover roughly the same area, meaning the the union of their spatial extents, $E^i$, is not empty, i.e. $\cup_{i=1}^t E^i \neq \emptyset$. Finally, we expect that these scenes are much larger than traditional image inputs in computer vision problems.  For example, scenes cover many square kilometers at sub-meter spatial resolution (i.e. each will have on the order of one hundred million pixels).

We would like to train a model that can \textit{count} and \textit{localize} cows/elk given new imagery of the same location, i.e. in the form $f(\mathbf{X}^{j}; \mathbf{\theta}) = \mathbf{\hat{Y}}^{j}$ where $\mathbf{\theta}$ are the parameters of the model and $j > t$. Here, $\mathbf{\hat{Y}}^{j}$ should be greater than $0$ where a cow or elk exists in $\mathbf{X}^{j}$ and $0$ otherwise. We fit the parameters $\theta$ following empirical risk minimization, for some loss function $J(f(\boldx; \theta), \boldy)$:
\begin{equation*}
    \min_\theta \frac{1}{t} \sum_{i=1}^t J(f(\boldx^t; \theta), \boldy^t)
\end{equation*}
In Section \ref{subsec:modeling} we discuss different architectural choices for $f(\cdot; \theta)$, as well as different loss functions for encoding the problem. Finally, we discuss metrics for measuring model performance in \textit{counting} and \textit{localizing} given a trained model and held out imagery/labels in Section \ref{subsec:metrics}.

\section{Methods}

\subsection{Modeling approaches} \label{subsec:modeling}

Following the methodology from density based crowd counting methods~\cite{lempitsky2010learning,boominathan2016crowdnet} we convert the binary point label masks, $\mathbf{Y}^i$, to either density masks, $\mathbf{Z}^i$, or segmentation masks, $\mathbf{S}^i$, to use in training of density based and localization based approaches respectively.

Specifically, to generate training labels for our density based approaches (CSRNet, FCRN, FCRN-LCN, and CowNet -- see below) we initialize a $7 \times 7$ Gaussian kernel, $K$, and convolve it with the given point label to produce a density mask, $\mathbf{Y}^i * K = \mathbf{Z}^i$. Here, we expect the sum of the predicted density will equal the number of predicted animals over different sized image regions. Similarly, for our segmentation approaches (UNet and CowNet -- see below) we apply a maximum filter of size $7 \times 7$ to $\mathbf{Y}^i$ and obtain a noisy mask of the objects, $\mathbf{S}^i$.

\begin{description}
\item[CowNet] is our proposed baseline approach. It is a UNet architecture, with two task-specific heads attached to the final spatial representation. The density head consists of a 1x1 convolution layer that predicts the per pixel density, while the segmentation head (also a 1x1 convolution layer) predicts the presence or absence of an animal. We train this architecture with a three term loss function designed to jointly optimize for: accurate segmentation, count density estimation, and global count. We train the segmentation head with labels $\mathbf{S}^i$, and the density head with labels \textbf{$\mathbf{Z}^i$}. 
\item[CSRNet~\cite{li2018csrnet}] is a density estimation-based neural network architecture for object counting. It uses a two VGG-like~\cite{simonyan2014very} convolutional neural networks (CNN) as a feature extractor and density estimator, respectively. Notably, the density estimator sub-network uses dilated kernel convolutions to deliver larger receptive fields and to replace all pooling layers. This approach is trained with a mean square error loss with labels \textbf{$\mathbf{Z}^i$}.
\item[LC-FCN~\cite{laradji2018blobs}] is a state of the art method for training segmentation based counting models (with previous applications to counting penguins, fish, cows and other objects~\cite{laradji2020counting,laradjiaffinity,laradji2018blobs}). Notably, the loss does not reason about the size or shape of the objects and instead forces the model to predict a blob of positive predictions for each object instance using point-level annotations. The loss consist of four terms: an image-level and a point-level loss, a split level loss to enforce the prediction of a single blob per object instance, and a false positive loss. We use the same architecture as~\cite{laradji2018blobs}, a FCN-18~\cite{long2015fully} with a ResNet18 backbone. The LC-FCN models are trained using \textbf{$\mathbf{Y}^i$}.
\item[FCRN~\cite{xie2018microscopy} / FCRN-LCN] is an architecture for detecting and counting cells in microscopy by performing density estimation. The FCRN-LCN variant includes local context normalization~\cite{ortiz2020local} instead of batch normalization layers. The network predicts object density at a pixel level and is trained with a \textbf{mean squared error loss} using \textbf{$\mathbf{Z}^i$} labels.
\item[UNet~\cite{ronneberger2015u}] is a popular encoder-decoder network architecture for performing general semantic segmentation tasks. UNets and various variants based on the UNet architecture have previously achieved state of the art results in a wide variety of application domains~\cite{falk2019u}. This approach is trained with a \textbf{pixel-wise weighted cross entropy loss} using \textbf{$\mathbf{S}^i$} labels.
\end{description}

\subsection{Measuring counting and localization performance with noisy point labels} \label{subsec:metrics}

Considering that we are interested in \textit{counting} and \textit{localizing} cows and elk in VHR satellite imagery \textit{scenes}, we need to measure performance with respect to the domain and purpose of the models -- large VHR satellite image scenes, and for ecological analyses. As 

\subsubsection{Counting} \label{subsubsec:counting-metrics}
With existing counting metrics such as mean absolute error (MAE) or mean absolute percentage error (MAPE), the difference between a predicted count, $\hat{y}$, and the ground truth count, $y$, is calculated per image and averaged over all images in a dataset. In our problem setting, we will have few large ``images'' (i.e. \textit{scenes}), therefore calculating MAE and MAPE will not be meaningful. The grid average mean absolute error (GAME) metric was introduced in~\cite{guerrero2015extremely} and extended to MAPE in~\cite{laradji2020counting} to measure how well counting algorithms also perform at localizing objects. GAME computes the average of the sum of the absolute errors calculated across non-overlapping windows from a dataset of $N$ images:
\begin{equation}
    \text{GAME}(L) = \frac{1}{N} \cdot \sum_{n=1}^{N} \left( \sum_{l=1}^{4^L} |\hat{y}_n^l - y_n^l | \right)
\end{equation}
where $L$ is a parameter that controls the number of non-overlapping windows that each image is divided into, and $\hat{y}_i^l$ is the $l^\text{th}$ window from the $i^\text{th}$ image in the dataset. When $L=0$, GAME is equivalent to MAE, while when $L>0$, GAME will measure how a model is able to localize an object with increasing fidelity.

We adapt this idea to work within a single \textit{scene} of VHR satellite imagery. We define a different version of ``gridded'' MAE, whereby a scene is split into a grid of $r \times r$ cells, $\mathcal{R}$, and the MAE is calculated over each cell:
\begin{equation}
    \text{GMAE}(r) = \frac{1}{|\mathcal{R}|} \sum_{i \in \mathcal{R}} |\hat{y}_i - y_i|.
\end{equation}
As satellite imagery has a fixed spatial resolution, $r$ is a physical distance. GMAE roughly measures how well a model is able to localize objects within a distance $r$ of their correct location. For example, if GMAE is high for small values of $r$, but lower for larger values of $r$, then the model is able to correctly identify how many objects there are in a scene, but is not able to correctly determine where each object is. This can be the case with ``glance'' type object counting methods~\cite{chattopadhyay2017counting} that are trained to directly regress the number of objects in a patch of input imagery, without necessarily localizing each object. To determine the per cell count for density based models, we sum the predicted density within each cell, while for segmentation based models, we sum the number of centroids of connected components of predicted positive pixels that fall within a cell. We can further normalize  $\text{GMAE}(r)$ to GMAE per km$^2$, assuming $r$ is in units of meters, by multiplying by $\frac{1000000}{r^2}$\footnote{Multiplying by $ \frac{1}{\text{m}^2\text{ per cell}} \cdot \frac{1000000\text{m}^2}{1\text{km}^2}$}. This is necessary to compare GMAE across different values of $r$ for a fixed area. 

Using the same grid setup as GMAE, we also compute the coefficient of determination (R$^2$) between model predicted counts and ground truth counts over all cells:
\begin{equation}
    \text{R}^2 = 1 - \frac{\sum_{i \in \mathcal{R}} (y_i - \hat{y}_i)^2}{\sum_{i \in \mathcal{R}} (y_i - \bar{y})^2}
\end{equation}
where $\bar{y} = \frac{1}{|\mathcal{R}|}\sum_{i \in \mathcal{R}} y_i$. Note that this value can be less than 0 when a model is making predictions that are worse than a hypothetical model which simply always returns the dataset mean.

\subsubsection{Localizing} \label{subsubsec:localization-metrics}

In \cite{laradji2018blobs} the authors point out that GAME might not be an accurate localization metric and compute classification metrics (specifically, F-Score) by matching predicted blobs to the point-level annotations. We extend this idea to our problem setting where point labels can be a distance $d$ away from the objects that they are annotating, and the objects themselves can be of a size less than $d^2$. Here, a reasonable model may (correctly) identify a group of pixels, that does not include the ground truth point label, as a positive class. Thus, we propose two matching algorithms for pairing predicted blobs with labeled points that takes this label uncertainty into account: an \textbf{optimistic} matching algorithm and a \textbf{conservative} matching algorithm.

In both algorithm we first group predicted foreground pixels in $\mathbf{\hat{Y}}^j$ into connected components (i.e. ``blobs'' in other object counting literature). In the \textbf{optimistic} algorithm we buffer each labeled point by a ``cutoff distance'' of $d$ and record the intersections between the buffered points and all connected components. The number of true positives is the number of labeled points that intersect with any predicted connected component, the number of false positives is the number of predicted connected components that do not intersect with any buffered points, and the number of false negatives is the number of buffered points that do not intersect with any predicted connected component. This is an ``optimistic'' accounting of a model's performance because a predicted connected component can intersect with more than one buffered label. In areas with dense labels (i.e. many cows/elk in a small space) this may be a desirable property as individual objects can be hard to separate. On the other hand, it will not penalize a model that over-segments the imagery in non-desirable cases. For example, a model that predicts a positive foreground class for an entire scene would not be penalized. This metric may be appropriate for ecological applications as finding cows/elk in large amounts of satellite imagery is similar to finding needles in a haystack. If \textit{any} predicted connected components are near to labeled points then the model has correctly localized the objects of interest, i.e. located the needles.

In the \textbf{conservative} algorithm we enforce the property that each predicted connected component can only count towards a single true positive. Here, we create a bipartite graph representation where each predicted connected component is represented as a node in the set $\mathbf{U}$ and each label is represented as a node in the set $\mathbf{V}$. We add edges between nodes $(u,v)$ if they are within the $d$ cutoff distance from each other. The number of true positives is now the size of the maximum cardinality matching on this graph, the number of false positives is the number of unmatched nodes from $\mathbf{U}$ and the number of false negatives is the number of unmatched nodes from $\mathbf{V}$. The maximum cardinality matching ensures that each predicted connected component is paired with a single labeled point. This accounting of a model's performance will correctly handle the degenerate case where a model predicts a positive foreground class for the entire input. Here, the ``optimistic'' algorithm will report perfect performance, while this algorithm will report a single true positive, no false positives, and $\text{``number of labels''} - 1$ false negatives.

\section{Data}
We use a dataset of 11 VHR panchromatic scenes from Maxar's satellite catalogue covering Point Reyes National Seashore, CA from 2013 to 2018. The spatial resolution of the scenes varies from $0.3 \text{ m}/\text{pixel}$ to $0.5 \text{ m}/\text{pixel}$, while off nadir position ranges from 11$^{\circ}$ to 34$^{\circ}$. The time of day and day of the year that each scene was imaged also differ between scenes, impacting lighting conditions and producing different shadow conditions throughout each image. We re-sample applicable scenes to a $0.3 \text{ m}/\text{pixel}$ resolution with bilinear interpolation. Multiple experts exhaustively annotated the positions of cattle and elk in each scenes, resulting in 10,529 labeled points.
See Figure \ref{fig:examples} to see a visualization of the panchromatic layer and labels from two areas of one of the test scenes.

\begin{table}[ht]
\centering
\begin{tabular}{@{}lcccc@{}}
\toprule
                & \multicolumn{2}{c}{\textbf{Localization Metrics}} & \multicolumn{2}{c}{\textbf{Counting Metrics}} \\ 
\cmidrule(lr){2-3}
\cmidrule(lr){4-5}
\textbf{Method} & Precision           & Recall              & GMAE               & R2               \\ \midrule
LC-FCN          & (0.563, 0.606)      & (0.779, 0.923)      & 0.134              & 0.766            \\
CowNet          & (0.465, 0.499)      & (0.787, 0.890)      & 0.241              & 0.557            \\
FCRN-LCN        & n/a                 & n/a                 & 0.190              & 0.212            \\
FCRN            & n/a                 & n/a                 & 0.268              & 0.038            \\
CSRNet          & n/a                 & n/a                 & 0.338              & 0.005            \\
UNet            & (0.387, 0.427)      & (0.605, 0.692)      & 0.304              & < 0              \\
\bottomrule
\end{tabular}%
\caption{Model performance averaged over three held out test scenes. The localization metrics are calculated with $d=4$ meters and shown as (conservative estimate, optimistic estimate). The counting metrics are calculated over an $r=100m$ grid covering the land area in each scene. The R$^2$ (i.e. the coefficient of determination) values can be less than $0$ -- this indicates worse performance than a model which simply predicts the dataset average ground truth count for each grid cell.}
\label{tab:merged}
\end{table}


\begin{figure*}[th]
\centering
\includegraphics[width=0.9\linewidth]{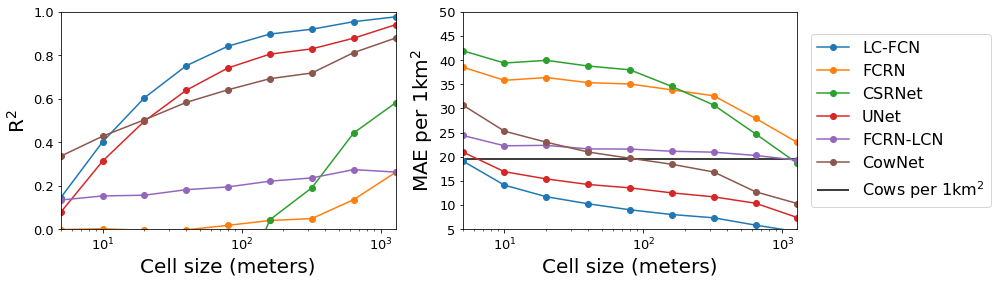}
\caption{Counting performance of different density based models on one held out scene. We tile the study area with different cell sizes and measure the GMAE/km$^2$ and R$^2$ between the predicted count of cows per grid cell and ground truth number of cows per cell for each tiling. The black horizontal line on the MAE/km$^2$ plots shows the ground truth density of cows ($\sim20/\text{km}^2$) recorded for the scene.}
\label{fig:quantitative_results}
\end{figure*}

\begin{figure*}[th]
\centering
\includegraphics[width=0.9\linewidth]{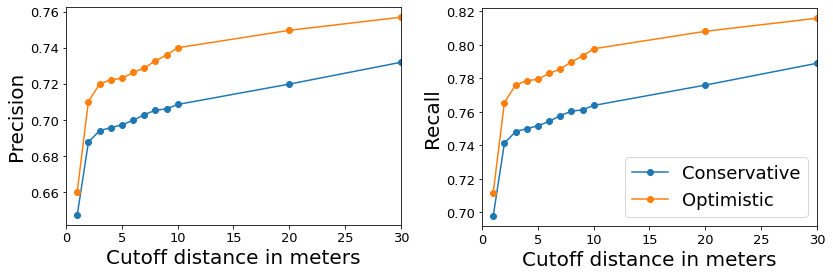}
\caption{We show the sensitivity of our method for computing precision and recall to the cutoff distance used with a set of model predictions over a held out scene. The blue line shows the \textit{conservative method} that enforces a one-to-one matching between connected components of predicted pixels and labeled points, while the orange line shows the \textit{optimistic method} that pairs any nearby predicted connected components with labeled points. We observe a sharp jump in estimated performance as the cutoff distance increases from 1 meter to 3 meters as the labeled points are not always precisely aligned to animals in the imagery, with diminishing returns after a cutoff of 4 meters.}
\label{fig:sensitivity}
\end{figure*}

\section{Experiments and Results}

To test the temporal generalization performance of the models, we select the earliest seven scenes from our dataset, spanning from February 2013 to December 2014, for training, a single scene from September 2016 for validation, and three scenes, spanning from March 2015 to April 2017, for testing. We train all models on the train split, select top performing models based on validation loss performance, and evaluate on the test split.

Table \ref{tab:merged} shows the average performance of all models over the test scenes. Here, the counting metrics over each scene are calculated at a $r=100$ meter resolution grid and the localization metrics are calculated with a cutoff distance of $d=4$. We evaluate both segmentation and density based models as described in Section \ref{subsubsec:counting-metrics}. The LC-FCN model performs the best across all metrics, for both counting and localization. We find that the LC-FCN, CowNet, and FCRN-LCN models are consistently making reasonable predictions across all scenes, while the performance of the other models is mixed. For example, the U-Net model suffers from many false positives/low precision, which reduces its counting performance metrics, and the CSRNet and FCRN models do not perform well in terms of counting metrics (i.e. estimating density) at a $100$ meter resolution. The LC-FCN and U-Net models have a large range between their conservative and optimistic recall estimates as they tend to predict large connected components of pixels for more dense groups of cattle which will only count as a single true positive in the conservative estimate.

We choose a cutoff distance of $d=4$ to compute the localization metrics based on the sensitivity analysis in Figure \ref{fig:sensitivity}. Here, we compute precision and recall of the CowNet model for different cutoff distances. We observe that there is a large jump in performance between $d=1$ and $d=2$ compared to subsequent values. This shows the effect of the label noise on the results (a significant number of labels are $>1$ meter away from the objects that they are annotating). As $d$ increases the estimated performance of the model increases as more predicted connected components are matched with labeled points that are farther away.  We choose $d=4$ as the cutoff to report values in Table \ref{tab:merged} as there is another jump in the ``optimistic'' matching performance between $d=4$ and $d=5$, and diminishing returns after. We believe this is a reasonable choice that reflects true model performance.

Figure \ref{fig:quantitative_results} shows the counting metrics as a function of grid cell size ($r$) for the density based models over one test scene. We observe that the performance of all the counting models improves with increasing cell size. With larger cell sizes, false positives and false negative errors can cancel out. At the lowest resolution grid cell size (1024 meters) the CowNet model achieves an 0.82 R$^2$. Notably, the R$^2$ if the CSRNet model is only positive after the grid cell size becomes larger than the training input size (i.e. when grid cells are larger than 256 pixels). CSRNet estimates coarse density over its inputs, therefore cannot localize at high resolutions.

\section{Conclusion}
In this work we benchmark different localization and counting based deep learning approaches for detecting cows and elk in very high-resolution satellite imagery. We specifically measure the temporal generalization of the models to test how they can perform in ecological monitoring settings when trained in a particular area. We propose evaluation methods that are tailored for this task, and find that an LC-FCN based models perform the best.

It is important for future work in this direction to a.) test the \textit{spatial generalization} of similar models; b.) compare the model estimated counts with ground truth counts or aerial survey results in order to determine how such models can fit into larger cattle monitoring efforts; c.) develop techniques for better training \textit{and evaluating} models in cases where there are dense groups of animals; and d.) develop post-processing methods appropriate for cleaning predictions made over VHR satellite imagery.

\bibliographystyle{ACM-Reference-Format}
\bibliography{citations}

\end{document}